\pdfoutput=1

\documentclass[10pt,twocolumn,letterpaper]{article}

\usepackage[pagenumbers]{cvpr} %

\usepackage{multirow}
\usepackage{makecell}
\usepackage{pifont}
\usepackage{algorithm}
\usepackage[noend]{algpseudocode}

\newcommand{\ours}{Unstick-Mem\xspace}  %
\newcommand{\oursI}{Unstick-ReInit\xspace}  %

\definecolor{upc}{RGB}{30,130,50}
\definecolor{dnc}{RGB}{200,40,40}
\newcommand{\sd}[1]{{\scriptsize$\pm$#1}}

\newcommand{\cmark}{\ding{51}}
\newcommand{\xmark}{\ding{55}}

\newcommand{\RM}{Mem\xspace}     %
\newcommand{\SP}{SP\xspace}     %
\newcommand{\OFT}{FT\xspace}   %
\newcommand{\RI}{ReInit\xspace}     %

\DeclareMathOperator*{\argmin}{arg\,min}

\definecolor{cvprblue}{rgb}{0.21,0.49,0.74}
\usepackage[pagebackref,breaklinks,colorlinks,allcolors=cvprblue]{hyperref}

\def\confName{CVPR}
\def\confYear{2027}

\title{Stuck in Place: Stall Diagnosis and Remedies for JEPA Planners}

\author{Junhyeok Lee\qquad Dongheon Lee\qquad Kyu Sung Choi\\
Seoul National University}

\begin{document}
\maketitle
\begin{abstract}
Joint-embedding predictive architecture (JEPA) world models plan from pixels with model predictive control, yet often fail on tasks their model should handle.
Test-time adaptation attributes such failures to model error under distribution shift and updates the weights on the current episode, without checking this attribution against the planning loop itself.
We follow the executed planning loop on pushing tasks.
In most settings, with and without distribution shift, nearly all failures (at least 96\%) are stalls, where the object stays in place while the model predicts progress.
The warm-started gradient-descent (GD) planner then returns nearly the same plan, a fixed point where the agent mostly idles.
We propose Unstick, which pairs on-policy fine-tuning before deployment with a fixed-weight loop breaker: a residual memory of the episode's prediction errors or random re-initialization at every replan.
With GD, Unstick recovers most of the adaptation gain on pushing tasks with the memory, and with re-initialization most of it on shape shifts but much less on visual shifts.
A large part of what limits these planners under GD is thus a fixed point of the planning loop, present with and without distribution shift, which a planner with fixed weights can break.
Code is available at \url{https://github.com/jhlee0619/unstick}.
\end{abstract}

\section{Introduction}
\label{sec:intro}

\begin{figure}[t]
  \centering
  \includegraphics[width=\columnwidth]{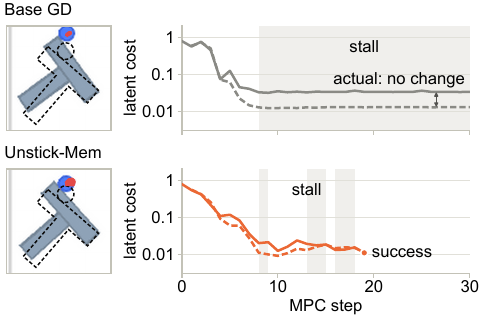}
  \caption{\textbf{Latent costs and frames of one PushObj-T episode.}
  Top: base GD planner; bottom: \ours (weights fixed at test time).
  Solid: actual latent cost; dashed: predicted after each chunk; arrow: predicted drop per chunk; gray band: stall.
  Frames: last step; blue: agent; red: agent path during the stall; dashed outline: goal.}
  \label{fig:teaser}
\end{figure}

A world model that predicts the consequences of actions lets an agent plan toward new goals from reward-free offline data~\cite{lecun2022path}.
Joint-embedding predictive architectures (JEPAs), which forecast the embedding of a future frame rather than its pixels, have become a standard route to planning from pixels toward goal images~\cite{zhou2025dinowm,sobal2025learning,assran2025vjepa2,wang2026straightening}.
With model predictive control (MPC)~\cite{garcia1989mpc}, the planner optimizes a short action sequence toward the goal embedding at every step, by gradient descent (GD) through the model~\cite{jyothir2023gradient,bharadhwaj2020model} or by the cross-entropy method (CEM)~\cite{rubinstein1997cem,hafner2019learning}, executes the first chunk, and replans.

When such a planner fails on a task its model should handle, test-time adaptation attributes the failure to model error under distribution shift and updates the weights on the current episode's transitions~\cite{sun2020ttt,wang2021tent,nagabandi2019learning}.
For JEPA world models, AdaJEPA~\cite{wang2026adajepa} does so after every executed chunk and reports large gains.
Analyses of latent planners likewise score plans or candidate actions~\cite{vesessook2026hidden,zhang2026arcbench,li2026where}.
Both views locate the failure in the model's predictions, and so expect that better predictions, if necessary from updated weights, make a better planner.
Neither view checks this attribution against the planning loop.

This expectation leaves out the loop, in which MPC replans from the states its own actions produce, so an error that leaves the state unchanged can recur at every replan, however well each plan is scored.
On pushing tasks, where an agent pushes a block to a goal pose, the executed loop of the \emph{base planner}, the pretrained model with standard MPC (GD unless stated otherwise), fails mostly through \emph{stalls}, in which the block stays in place while the model predicts progress (\cref{fig:teaser}).
In most settings, nearly all failures (at least 96\%) are stalls, including those on the training shapes and clean frames, so shift alone does not explain them.
Test-time adaptation also adds a backward pass to every replan and risks forgetting and collapse~\cite{niu2022efficient,niu2023towards}.
We therefore trace what keeps the loop in place and test whether a planner with fixed weights can break it.

Two ingredients keep the loop in place.
At the stalled states we analyze, the model, not the search, is mostly at fault: it never prefers a one-chunk plan that would succeed to staying, whereas the objective on simulated frames often does.
The executed chunk moves the agent out and back. The deterministic GD planner, warm-started from its previous plan as in real-time MPC~\cite{diehl2005real}, then returns nearly the same plan from the unchanged state, so the loop reaches a \emph{fixed point} in which the agent mostly idles.
With CEM, which samples new plans, the agent moves, yet the block stays in place.

These two ingredients hold the planning loop at the fixed point, which neither more accurate predictions alone nor test-time weight updates target directly.
We therefore propose \emph{Unstick}, which pairs \emph{on-policy fine-tuning} (\OFT) with a fixed-weight loop breaker (\cref{fig:teaser}).
Fine-tuning corrects the model, before deployment, on the base planner's own executions of separate episodes, because the planner visits states that offline data miss~\cite{parthasarathy2025closing,ross2012agnostic}, but alone leaves the agent idle in most remaining failures.
In \ours, a \emph{residual memory} (\RM) adds the model's one-step errors on executed chunks back to predictions near them, so a re-proposed stay-in-place chunk is predicted to stay in place.
In \oursI, \emph{random re-initialization} (\RI) starts every replan from a random plan instead of the warm start, so the same state no longer yields the same plan.
With GD and after on-policy fine-tuning, the memory recovers most of the gain of test-time adaptation on pushing tasks, and re-initialization most of it on shape shifts but much less on visual shifts.
A large part of what limits these planners under GD is thus a fixed point of the planning loop, not an error that only test-time weight updates can fix.
JEPA planners should therefore be evaluated in closed loop, by checking whether the state moves and whether the plan repeats, before their failures are attributed to distribution shift, and test-time adaptation should be compared with remedies that keep the weights fixed.

\noindent Our contributions are:
\begin{itemize}
\item Closed-loop diagnostics show that JEPA planners on pushing tasks fail mainly by stalling, which a deterministic GD planner turns into a fixed point.
\item We break the stall without test-time weight updates, with an episodic memory of model residuals or random re-initialization, which after on-policy fine-tuning recover most of the adaptation gain.
\item We trace failure cases that neither remedy fixes to an encoder that discards a task variable or to local minima of the planning objective, rather than to the planner.
\end{itemize}
\section{Related work}
\label{sec:related}
\begin{figure*}[t]
  \centering
  \includegraphics[width=\textwidth]{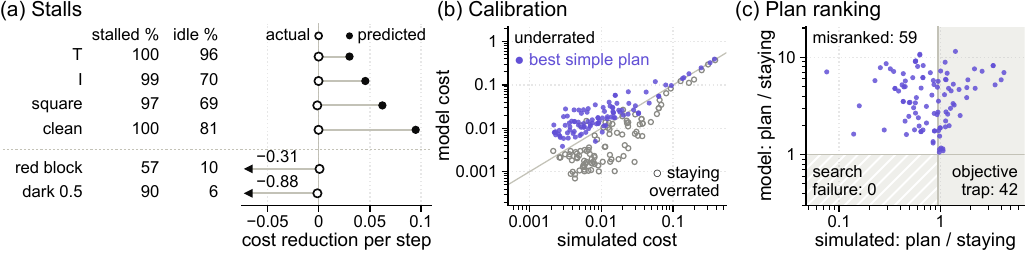}
  \caption{\textbf{Failure analysis of the base GD planner.}
  (a) Idle: agent moves $<$2\,px per MPC step. (b, c) The 101 stalls at MPC step 25 (T, clean); (c) labels give counts.
  Objective trap: no simple plan lowers the simulated cost of staying by $\geq$5\%; misranked: a simple plan lowers the simulated cost, but the model prefers staying.}
  \label{fig:analysis}
\end{figure*}

\paragraph{Latent planning.}
Planning from pixels with learned models goes back to video prediction for robotic pushing~\cite{finn2017deep} and latent dynamics planned with CEM~\cite{hafner2019learning}, and now covers manipulation and navigation~\cite{assran2025vjepa2,bar2025navigation}.
JEPAs predict future latents rather than pixels~\cite{lecun2022path,zhou2025dinowm,sobal2025learning,wang2026straightening}.
Interleaved CEM and gradient steps~\cite{bharadhwaj2020model} and stochastic relaxed state trajectories~\cite{psenka2026parallel} counter local optima of gradient-based planning, warm starts give fast feedback in real-time MPC~\cite{diehl2005real}, and PlaNet restarts CEM after every observation~\cite{hafner2019learning}.
We study the closed loop across replans, where a model error and the warm-started GD planner form a stay-in-place fixed point.

\paragraph{Failure analyses.}
Analyses of latent planners show that scoring plans at the end of the horizon can misjudge model quality~\cite{vesessook2026hidden}, and that frozen JEPA world models misrank candidate actions, which frequent replanning masks~\cite{zhang2026arcbench}. Another analysis measures how much prediction, the decision metric, and search each cost~\cite{li2026where}.
Other work searches for failure-inducing inputs~\cite{shi2026where} or audits imagined interventions~\cite{vakalis2026intervention}.
These analyses score plans, candidate actions, or open-loop predictions, whereas we follow the closed loop, find that it stalls, and break the stall with fixed weights.

\paragraph{Test-time adaptation.}
Test-time methods update a pretrained model on unlabeled test inputs to counter distribution shift~\cite{sun2020ttt,wang2021tent}, in control also by meta-learning inside MPC~\cite{nagabandi2019learning}.
AdaJEPA~\cite{wang2026adajepa} brings this to JEPA world models with a one-step update per replan.
JEPA-TTT~\cite{zhang2026jepattt} keeps training the predictor across test episodes under dynamics shifts, and Sandwich-Residuals~\cite{soni2026sandwich} optimizes small residual modules around a frozen predictor online; our memory stores observed residuals and optimizes nothing at test time.
WM-LAN~\cite{bhattarai2026wmlan} infers an unobserved dynamics context by searching the latent code of an encoder--decoder world model; our memory corrects predictions directly near visited transitions.

\paragraph{Model exploitation.}
A planner that optimizes against a learned model seeks out the model's errors, which model-based RL counters with uncertainty-aware models~\cite{chua2018pets}, short rollouts~\cite{janner2019mbpo}, and control-matched objectives~\cite{lambert2020objective}.
DAgger and agnostic system identification iteratively refit on data from the current controller~\cite{ross2011dagger,ross2012agnostic}, and Agrawal~\etal~\cite{parthasarathy2025closing} fine-tune on planner-induced or adversarial trajectories.
We fine-tune on a single round of base-planner executions and pair the fine-tuning with a loop breaker.

\paragraph{Residual models and episodic memory.}
In learning-based MPC for racing, Gaussian processes fit the error of a nominal vehicle model~\cite{hewing2020cautious}, also online~\cite{kabzan2019learning}, and in RL, episodic memories estimate values~\cite{pritzel2017nec}.
Our memory corrects a learned visual world model in its latent space and is rebuilt in every episode.

\paragraph{Degeneration in language models.}
Deterministic decoding from language models produces repetitive, degenerate text~\cite{holtzman2020curious}, remedied by repetition penalties~\cite{keskar2019ctrl}, sampling~\cite{holtzman2020curious}, or unlikelihood training on the model's own outputs~\cite{welleck2020unlikelihood}.
Language models exploit the current sequence by updating the weights on it (dynamic evaluation~\cite{krause2018dynamic}, the counterpart of test-time adaptation) or by storing it and adjusting predictions without updates (cache and nearest-neighbor models~\cite{grave2017cache,khandelwal2020knnlm}).
Latent MPC has a similar loop, deterministic and with a model that favors a degenerate action; our residual memory and random re-initialization break it with fixed weights.

\section{Stall diagnosis}
\label{sec:analysis}

\subsection{JEPA planning}
\label{sec:prelim}

A JEPA world model consists of an observation encoder $\mathcal{E}_\phi$, an action encoder $\mathcal{E}_\psi$, and a predictor $f_\theta$:
\begin{equation}
  z_t = \mathcal{E}_\phi(o_t), \quad u_t = \mathcal{E}_\psi(a_t), \quad \hat z_{t+1} = f_\theta(z_{t-h:t}, u_{t-h:t}),
\end{equation}
where $a_t$ is the action chunk applied between observations $o_t$ and $o_{t+1}$, and $h$ is a short history window.
Below we abbreviate $f_\theta(z_{t-h:t}, u_{t-h:t})$ as $f_\theta(z_t, \mathcal{E}_\psi(a_t))$, with the history implied.
The model is trained on reward-free offline trajectories with a latent prediction loss $\ell(\hat z_{t+1}, \mathrm{sg}(z_{t+1}))$, where $\mathrm{sg}$ is the stop-gradient operator.
Given a goal image with latent $z_g$, MPC with horizon $H$ solves
\begin{equation}
  a^\star_{t:t+H-1} = \argmin_{a_{t:t+H-1}} \; J(a) = \sum_{k=1}^{H} \alpha_k \, \lVert \hat z_{t+k} - z_g \rVert_2^2,
  \label{eq:mpc}
\end{equation}
executes the first chunk $a_t$, observes $o_{t+1}$, and replans; one such cycle is an \emph{MPC step}.
The default weights $\alpha_k$ emphasize the end of the horizon, with about 3\% of the weight on the frame reached by the executed chunk and 51\% on the last frame.
We solve \cref{eq:mpc} either with GD, warm-started from the unexecuted tail of the previous plan, or with CEM~\cite{rubinstein1997cem}.

\subsection{Stall statistics}
\label{sec:stall}
We analyze the base planner with GD on PushObj~\cite{zhou2025dinowm} (seen T, unseen I and square) and PushT~\cite{chi2023diffusion} (clean frames and visual shifts: blur, salt-and-pepper noise (snp), darkening, and red recolorings of the agent, goal region, or block), replaying executed actions in the simulator to recover true states.
We label an episode \emph{stalled} if, in the simulator, the block moves less than 5\,px and 5$^\circ$ over MPC steps 20--30.
We call the agent \emph{idle} if it moves less than 2\,px per MPC step over the same steps.
On T, I, square, and clean, at least 97\% of the failures are stalls, and the agent is idle in 69--96\% of them.
During these stalls, the model predicts progress (a reduction of the planning cost) of 0.030--0.095 per executed chunk, whereas the actual progress is zero (\cref{fig:analysis}a).
On T, the predicted progress of 0.030 per chunk exceeds the median final cost of successful episodes (0.021).
Over all 15 pushing settings (seven shapes and eight visual conditions, including darkening at level 0.5), 12 reach at least 96\% stalls, and red goal reaches 93\%. In 10 of these 12, the agent is idle in 69--96\% of the failures (19\% under snp, 55\% under red agent).
Failures under the red block and darkening at level 0.5, where the predicted progress is negative, and idle failures in PointMaze-Medium~\cite{fu2020d4rl} differ from these stalls. Their causes lie in the encoder (red block, darkening) and in a local minimum of the planning objective (PointMaze-Medium).

\subsection{Plan ranking at stalled states}
\label{sec:diagnosis}
At MPC step 25 of every failure whose agent is idle over MPC steps 20--25 (64 on T, 37 on clean), we score the planner's own plan, 51 simple plans, and 8 plans reoptimized from random initializations.
A simple plan executes one straight chunk and then holds.
Each plan is scored by the planning objective under the model and on the frames produced in the simulator (\cref{fig:analysis}b,c).
The model overrates (underrates) a plan if it assigns the plan a lower (higher) cost than the simulated frames do.
\textbf{Search.}
Within the plans we score, search is not the bottleneck.
Re-initialized plans find a lower model cost than the planner's own plan in only 5--11\% of stalls.
We call a stall a search failure if a simple plan beats staying on simulated frames, is preferred by the model, and is missed by the planner.
No stall is a search failure (\cref{fig:analysis}c).
\textbf{Calibration.}
On T, the model overrates staying in 98\% of stalls and underrates the best simple plan (median relative error of the model cost against the simulated cost, $+92\%$; \cref{fig:analysis}b).
Simple plans move and then hold, which is rare in expert data, so the share of underrated moving plans may be inflated.
The model never prefers a simple plan that would succeed, whereas the objective on simulated frames does in 58\% (T) and 77\% (clean) of the stalls where such a plan exists.
In 45\% of stalls on T, no simple plan lowers the simulated cost of staying by at least 5\%, and we call these stalls \emph{objective traps} within our plan set.
The model overrates staying in 97\% of them as well.

\subsection{Warm-start fixed point}
\label{sec:fixedpoint}
The executed chunks move the agent back and forth, with a net displacement below 0.3 of the path length for 89--99\% of GD chunks against 3\% of expert chunks on T.
Because the agent comes back to its start, the state does not change, and the warm-started planner returns nearly the same plan at the next step (median relative change of 0.07--0.08 between consecutive chunks of failed episodes on T, I, and clean).
The loop has reached a fixed point, and the base planner's success stops rising after about 10 MPC steps (\cref{fig:curves}).
The fixed point needs the deterministic planner, but the stall does not.
With CEM, the median relative change between consecutive chunks is 1.1--1.3 and no agent idles, yet the block stays in place in 90--99\% of the failures on T, I, and clean, while the model predicts progress that does not occur.
Neither straight chunks, which replace each chunk's actions by their mean, nor uniform weights on all predicted frames of the objective remove the stall (supplementary material).

\section{Stall remedies}
\label{sec:method}
The diagnosis gives two targets: the model overrates staying at the analyzed stalls, and warm-started GD turns this error into a fixed point.
A different search alone does not remove the stall, since with CEM the block also stays in place in most failures.
The residual memory addresses the first target at planning time, random re-initialization the second, and on-policy fine-tuning the first before deployment.

\subsection{Residual memory}
\label{sec:rm}
On the model side, a \emph{residual memory} (\RM) keeps a stay-in-place chunk from looking like progress by correcting the predictions near the episode's own transitions, as a cache model does~\cite{grave2017cache,khandelwal2020knnlm}.
Each episode keeps a memory $\mathcal{M}$ of its executed chunks.
After executing chunk $a_t$, we store the key $(z_t, a_t)$ and the model's one-step error on this transition,
\begin{equation}
  r_t = z_{t+1} - f_\theta\big(z_t, \mathcal{E}_\psi(a_t)\big),
  \label{eq:res}
\end{equation}
where $z_t$ and $z_{t+1}$ encode the observations before and after the chunk, and $f_\theta$ uses the same $h$-frame context as the planner's rollout.
During planning, each predicted step $\hat z_{k+1} = f_\theta(\hat z_k, \mathcal{E}_\psi(a_k))$ of each candidate plan is corrected with the residuals stored near its input:
\begin{gather}
  \hat z_{k+1} \leftarrow \hat z_{k+1} + \frac{\sum_{i\in\mathcal{M}} w_i\, r_i}{\max\big(1, \sum_{i\in\mathcal{M}} w_i\big)},
  \label{eq:rm}\\
  w_i = \exp\Big(-\frac{\lVert a_k - a_i\rVert_2^2}{2\sigma_a^2} - \frac{d(\hat z_k, z_i)^2}{2\sigma_z^2}\Big),
  \label{eq:w}
\end{gather}
where actions are normalized and $d$ is the root-mean-square distance between latents.
Far from all stored transitions, the kernel weights $w_i$ vanish, and the $\max(1,\cdot)$ keeps a single distant residual from being extrapolated.
Residuals are constants, and GD receives gradients through the model and the kernel weights.
We set $\sigma_a=1$ and choose $\sigma_z$ per model by a fixed rule: the median distance $d$ between consecutive observed latents in executions of the base planner on the fine-tuning episodes.

If the planner re-proposes a chunk that it executed from the same state, the corrected rollout predicts what was observed, so the predicted progress that kept the loop on this chunk disappears.
The memory adds a one-step prediction per MPC step and a kernel sum over $\mathcal{M}$ per predicted step.

\subsection{Random re-initialization}
\label{sec:ri}
On the planner side, \emph{random re-initialization} (\RI) removes the determinism of the warm-started GD planner, which returns the same plan from the same state.
Re-initialization instead starts every replan after the first from a plan drawn from $\mathcal{N}(0, I)$ in the normalized action space and runs the same 100 GD iterations.
Random restarts are common in sampling-based planners~\cite{hafner2019learning}; here they break the warm-start fixed point of GD.
Any executed chunk that moves the agent then changes the state.
Re-initialized plans rarely find a lower model cost at a stall, so re-initialization leaves the ranking uncorrected and removes only the deterministic loop.
It adds no computation.

\subsection{On-policy fine-tuning}
The residual memory corrects the model only where the current episode has been.
\emph{On-policy fine-tuning} (\OFT) on the planner's own outputs~\cite{welleck2020unlikelihood} corrects, before deployment, the model's errors on the chunks that the planner produces, including those that recur across episodes.
We run the base planner on \emph{fine-tuning episodes}, whose start states differ from those of the evaluation episodes, and record its executions $\mathcal{D}_\text{on}$, stalls included.
On a set $\mathcal{B}$ of executed transitions, test-time adaptation minimizes the one-step latent prediction loss
\begin{equation}
  \mathcal{L}_\text{ada}(\mathcal{B}) = \frac{1}{|\mathcal{B}|}\sum_{(o_i,a_i,o_{i+1})\in\mathcal{B}} \ell\big(f_\theta(z_i, \mathcal{E}_\psi(a_i)),\, \mathrm{sg}(z_{i+1})\big).
  \label{eq:ada}
\end{equation}
We minimize it over sliding windows of $\mathcal{D}_\text{on}$, on the parameters $\Omega$ that adaptation updates (the last predictor block with the final LayerNorm, plus the encoder projection head) with adaptation's learning rates.
The data come from a single round of simulator executions, without expert actions, rewards, success labels, or the original training set.
Nothing beyond the model's own inputs is needed at deployment.
In \ours and \oursI, the deployed weights stay fixed, no parameter gradients are computed at test time, and all episodes share one model.

\begin{table*}[t]
\centering
\caption{\textbf{Success on seen and unseen shapes (PushObj).} GD, success (\%). \ours: on-policy fine-tuning (\OFT) + residual memory (\RM); \oursI: \OFT + random re-initialization (\RI); AdaJEPA: test-time weight updates. Mean $\pm$ SD over seeds. \textbf{Bold}, \underline{underline}: best, second best without test-time weight updates; ties share marks.}
\label{tab:shape}
\small
\setlength{\tabcolsep}{3.4pt}
\begin{tabular}{l ccccccc}
\toprule
Success (\%) $\uparrow$ & \multicolumn{4}{c}{Seen shapes} & \multicolumn{3}{c}{Unseen shapes} \\
\cmidrule(lr){2-5}\cmidrule(lr){6-8}
Method & T & L & Z & cross & I & small T & square \\
\midrule
Base                                       & 53.3\sd{6.8} & 50.7\sd{6.6} & 44.7\sd{5.0} & 52.0\sd{4.3} & 26.0\sd{5.7} & 42.7\sd{5.2} & 23.3\sd{3.4} \\
\quad + \ours                              & \textbf{78.7}\sd{5.7} & \textbf{82.0}\sd{4.3} & \textbf{78.7}\sd{3.8} & \textbf{77.3}\sd{3.8} & \underline{40.7}\sd{1.9} & \textbf{68.7}\sd{3.8} & \underline{36.0}\sd{3.3} \\
\quad + \oursI                             & \underline{76.0}\sd{4.3} & \underline{71.3}\sd{2.5} & \underline{76.7}\sd{1.9} & \underline{71.3}\sd{5.7} & \textbf{50.0}\sd{11.8} & \underline{68.0}\sd{5.9} & \textbf{42.7}\sd{0.9} \\
\quad + straight chunks                    & 55.3\sd{0.9} & 49.3\sd{1.9} & 45.3\sd{2.5} & 48.0\sd{4.9} & 22.0\sd{4.3} & 42.0\sd{14.2} & 22.7\sd{3.8} \\
\quad + uniform weights                    & 60.0\sd{3.3} & 57.3\sd{4.1} & 56.7\sd{2.5} & 64.7\sd{6.6} & 33.3\sd{8.1} & 58.0\sd{5.9} & 20.7\sd{0.9} \\
\midrule
AdaJEPA~\cite{wang2026adajepa} & 84.7\sd{5.7} & 86.7\sd{3.8} & 79.3\sd{5.0} & 78.7\sd{1.9} & 48.0\sd{8.6} & 65.3\sd{3.4} & 36.7\sd{2.5} \\
\bottomrule
\end{tabular}
\end{table*}

\section{Experiments}
\label{sec:experiments}
\begin{figure*}[t]
  \centering
  \includegraphics[width=\textwidth]{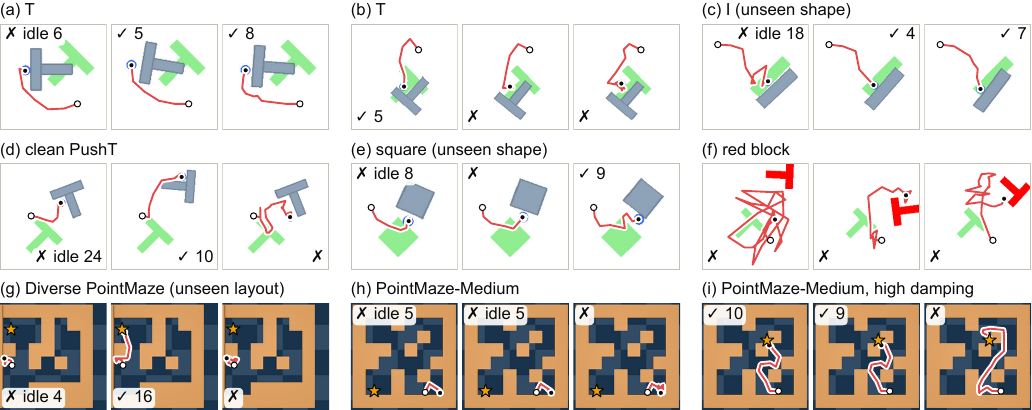}
  \caption{\textbf{First episode per outcome (GD, seed 1).}
  Each triplet: base planner, \ours, \oursI (PointMaze: base planner, residual memory, random re-initialization). Last frame; agent path, start ($\circ$) to last step ($\bullet$); green: block goal; star: maze goal; \cmark\,$k$: success at MPC step $k$; \xmark\ idle $k$: idle from MPC step $k$ on; \xmark: fails.}
  \label{fig:qual}
\end{figure*}

\subsection{Setup}
\label{sec:setup}
\noindent\textbf{Benchmarks.}
We use the public JEPA world models of Wang~\etal~\cite{wang2026adajepa} (our main models), all trained with the temporal-straightening loss~\cite{wang2026straightening}, and their evaluation sets.
\emph{PushObj (seen and unseen shapes):} trained on block shapes $\{$T, L, Z, cross$\}$, tested on these and on the unseen $\{$I, small T, square$\}$.
\emph{PushT (visual shifts):} trained on clean renderings, tested on clean frames, blur, 1\% salt-and-pepper noise (snp), darkening (level 0.9), and red recolorings of the agent, goal region, and block.
\emph{PointMaze (dynamics and layout shifts):} PointMaze-Medium (medium maze) with default dynamics, low mass ($\times0.2$), and high damping ($\times20$), and Diverse PointMaze~\cite{sobal2025learning,zhang2026hierarchical} on five unseen layouts.

\noindent\textbf{Planning and evaluation.}
The horizon is $H=5$ chunks (25 environment steps; frameskip 5), one chunk of 5 actions is executed per MPC step, and the MPC budget is 30 steps on PushT and PushObj and 20 on PointMaze.
GD and CEM follow~\cite{wang2026adajepa}.
A PushT episode succeeds when the joint position error of the agent and the block is below 20\,px and the block's angle error is below 20$^\circ$.
We report mean $\pm$ standard deviation (SD) over three evaluation seeds of 50 episodes.
We compare in detail on four settings: the seen shape T, the unseen shapes I and square, and clean PushT. The four-setting average is the unweighted mean of their cells.
Further checks use 150 fresh episodes per setting, disjoint from all other episodes, and two models with frozen DINOv2~\cite{oquab2024dinov2} encoders: DINO-WM~\cite{zhou2025dinowm} and Temporal Straightening~\cite{wang2026straightening}.

\noindent\textbf{Settings of our methods.}
On-policy data cover seen shapes only for the PushObj model (379 executions) and clean renderings only for the PushT model (89), so fine-tuning sees neither unseen shapes nor visual shifts.
Fine-tuning is kept only where adding it to the stall penalty does not lower success on a separate selection set of episodes in the model's training setting.
On PointMaze with GD it is dropped, so we report the memory and re-initialization alone, with fine-tuned rows for comparison.

\noindent\textbf{Baselines and upper reference.}
We compare with methods that also keep the weights fixed at test time: the base planner, fine-tuning alone, and two planning-time baselines that use a stall flag.
After each executed chunk $a_t$, a stall is flagged, $s_t = \mathbf{1}[\lVert p_{t+1} - p_t \rVert_2 < \delta] \wedge \neg\,\text{success}_t$, when the agent's proprioceptive position $p$, observable at test time, has moved less than $\delta$ ($2$\,px on the pushing tasks, the idle threshold) and the episode has not succeeded.
The memory, re-initialization, and adaptation use neither the agent position nor the success signal.
\emph{Restart at stalls} re-initializes after a flagged stall.
A \emph{stall penalty} (\SP)~\cite{keskar2019ctrl} adds to $J$ a Gaussian penalty around the chunks that ended in a stall.
The upper reference is AdaJEPA~\cite{wang2026adajepa}, which after each chunk takes one gradient step of $\mathcal{L}_\text{ada}$ on $\Omega$ over the episode's five most recent transitions, starting each episode from the pretrained weights.
On one GPU, an MPC step for the 50 episodes of a seed on PushObj-T takes 7.4\,s with \ours and 6.8\,s with \oursI, against 6.7\,s for the base planner and 8.3\,s for AdaJEPA, which keeps weights per episode.
Peak GPU memory is 1.5\,GiB with \ours and 0.9\,GiB with \oursI, against 0.9\,GiB for the base planner and 1.6\,GiB for AdaJEPA.
All methods share world models, planners, seeds, and episodes.
Model, protocol, hyperparameter, selection, and baseline details are in the supplementary material.

\begin{table*}[t]
\centering
\caption{\textbf{Success under visual shifts (PushT).} GD, success (\%). \ours: on-policy fine-tuning (\OFT) + residual memory (\RM); \oursI: \OFT + random re-initialization (\RI); AdaJEPA: test-time weight updates; snp: salt-and-pepper noise; dark: darkening level. Mean $\pm$ SD over seeds. \textbf{Bold}, \underline{underline}: best, second best without test-time weight updates; ties share marks.}
\label{tab:visual}
\small
\setlength{\tabcolsep}{3.4pt}
\begin{tabular}{l ccccccc}
\toprule
Success (\%) $\uparrow$ & No shift & \multicolumn{6}{c}{Visual shifts} \\
\cmidrule(lr){2-2}\cmidrule(lr){3-8}
Method & clean & blur & snp & dark 0.9 & red agent & red goal & red block \\
\midrule
Base                                       & 64.0\sd{1.6} & 56.0\sd{3.3} & 64.7\sd{1.9} & 63.3\sd{1.9} & 44.0\sd{9.1} & 34.7\sd{5.2} & \underline{11.3}\sd{2.5} \\
\quad + \ours                              & \textbf{80.7}\sd{0.9} & \textbf{76.7}\sd{2.5} & \textbf{74.0}\sd{1.6} & \textbf{78.0}\sd{4.3} & \textbf{62.0}\sd{5.9} & 38.7\sd{5.0} & \underline{11.3}\sd{2.5} \\
\quad + \oursI                             & \underline{69.3}\sd{6.6} & 59.3\sd{2.5} & 68.7\sd{6.6} & \underline{68.0}\sd{5.9} & \underline{58.7}\sd{11.8} & \underline{40.0}\sd{5.9} & 9.3\sd{0.9} \\
\quad + straight chunks                    & 57.3\sd{3.4} & 40.0\sd{4.3} & 59.3\sd{5.2} & 55.3\sd{9.8} & 47.3\sd{5.2} & 36.0\sd{3.3} & \textbf{12.0}\sd{4.9} \\
\quad + uniform weights                    & 66.7\sd{2.5} & \underline{66.0}\sd{1.6} & \underline{69.3}\sd{0.9} & 66.7\sd{5.0} & 56.0\sd{4.9} & \textbf{44.0}\sd{7.1} & 9.3\sd{1.9} \\
\midrule
AdaJEPA~\cite{wang2026adajepa} & 84.0\sd{1.6} & 86.0\sd{1.6} & 82.7\sd{3.4} & 83.3\sd{1.9} & 59.3\sd{6.2} & 49.3\sd{5.2} & 8.7\sd{2.5} \\
\bottomrule
\end{tabular}
\end{table*}
\subsection{Comparison with test-time adaptation}
\label{sec:main}
\begin{figure}[t]
  \centering
  \includegraphics[width=\columnwidth]{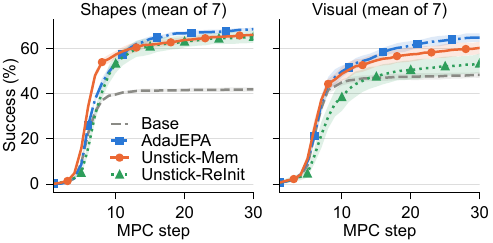}
  \caption{\textbf{Success over MPC steps.} GD; averages over the seven shapes (left) and the seven visual conditions (right). Lines: mean over 3 seeds; shading: SD.}
  \label{fig:curves}
\end{figure}
\begin{table*}[t]
\centering
\caption{\textbf{Component analysis.} GD, success (\%). \OFT: on-policy fine-tuning; Plan.: planning-time change; \SP: stall penalty; \RM: residual memory; \RI: random re-initialization; AdaJEPA: test-time weight updates; $^\star$: unseen. Mean $\pm$ SD over seeds. \textbf{Bold}, \underline{underline}: best, second best without test-time weight updates; ties share marks.}
\label{tab:ablation}
\footnotesize
\setlength{\tabcolsep}{3.6pt}
\begin{tabular}{l cc ccccc}
\toprule
Method & \OFT & Plan. & T $\uparrow$ & I$^\star$ $\uparrow$ & square$^\star$ $\uparrow$ & clean $\uparrow$ & red block$^\star$ $\uparrow$ \\
\midrule
Base                                       &        &                        & 53.3\sd{6.8} & 26.0\sd{5.7} & 23.3\sd{3.4} & 64.0\sd{1.6} & 11.3\sd{2.5} \\
\quad + restart at stalls                  &        & {\scriptsize restart}  & 63.3\sd{5.7} & 31.3\sd{5.2} & 26.0\sd{4.3} & 70.0\sd{2.8} & 12.7\sd{4.1} \\
\quad + \SP                                &        & \SP                    & 64.7\sd{5.7} & 36.7\sd{6.8} & 30.0\sd{4.3} & 71.3\sd{0.9} & 12.0\sd{3.3} \\
\quad + \RM                                &        & \RM                    & 68.7\sd{9.0} & 31.3\sd{5.7} & 28.7\sd{3.4} & \underline{76.7}\sd{4.1} & 10.7\sd{3.4} \\
\quad + \RI                                &        & \RI                    & 66.0\sd{4.3} & \underline{44.7}\sd{3.4} & 35.3\sd{3.4} & 68.7\sd{4.1} & \textbf{14.7}\sd{2.5} \\
\quad + \RI + \RM                          &        & \RI, \RM               & 66.0\sd{4.3} & 44.0\sd{2.8} & 35.3\sd{3.4} & 68.7\sd{4.1} & 12.7\sd{2.5} \\
\quad + \OFT                               & \cmark &                        & 68.0\sd{9.9} & 34.7\sd{1.9} & 31.3\sd{5.0} & 66.0\sd{1.6} & 12.7\sd{2.5} \\
\quad + \ours (\OFT + \RM)                 & \cmark & \RM                    & \textbf{78.7}\sd{5.7} & 40.7\sd{1.9} & \underline{36.0}\sd{3.3} & \textbf{80.7}\sd{0.9} & 11.3\sd{2.5} \\
\quad + \oursI (\OFT + \RI)                & \cmark & \RI                    & 76.0\sd{4.3} & \textbf{50.0}\sd{11.8} & \textbf{42.7}\sd{0.9} & 69.3\sd{6.6} & 9.3\sd{0.9} \\
\quad + \OFT + \RI + \RM                   & \cmark & \RI, \RM               & \underline{76.7}\sd{4.7} & \textbf{50.0}\sd{11.8} & \textbf{42.7}\sd{0.9} & 71.3\sd{5.2} & \underline{14.0}\sd{1.6} \\
\midrule
AdaJEPA                                    &        &                        & 84.7\sd{5.7} & 48.0\sd{8.6} & 36.7\sd{2.5} & 84.0\sd{1.6} & 8.7\sd{2.5} \\
\bottomrule
\end{tabular}
\end{table*}
\noindent\textbf{Shape and visual shifts.}
With GD, \ours and \oursI recover 80\% and 82\% of the adaptation gain on the four settings (mean gain over the base planner relative to AdaJEPA's), 91\% and 88\% on the seven shapes, and 72\% and 31\% on the seven visual conditions (\cref{tab:shape,tab:visual}). Over all 14 settings, \ours recovers 84\%.
Each beats the base planner in 13 of the 14 settings, all but the red block.
Under visual shifts, fine-tuning has seen only clean renderings, so the memory, whose residuals come from the episode's own shifted frames, is the only component that corrects the model with them.
Re-initialization leaves the model unchanged, which matches its smaller share of the gain there.
With both remedies, success keeps rising after that of the base planner stops (\cref{fig:curves}).
\Cref{fig:qual} shows rescued (a, c--e, g), lost (b, i), and unchanged (f, h) episodes.

\noindent\textbf{\ours versus \oursI.}
\oursI is ahead of \ours on two unseen shapes, I and square (\cref{fig:qual}e), and behind on T and clean (\cref{tab:shape,tab:visual}).
On all three unseen shapes, \oursI exceeds adaptation, so adaptation is an upper reference, not a bound.

\noindent\textbf{Fresh episodes.}
Relative to the base planner, \ours rescues 134 of 600 episodes (fail with the base planner, succeed with \ours) and loses 41 (the reverse), and \oursI rescues 161 and loses 54.
The memory alone raises success on T, I, square, and clean without losing an episode (supplementary material).

\subsection{Component analysis}
\label{sec:ablation}

\noindent\textbf{Memory and fine-tuning.}
The memory and fine-tuning correct different errors (\cref{tab:ablation}).
Fine-tuning reduces the number of failures (\eg, 70 $\to$ 48 on T), but the agent is still idle in 94\% of them.
The memory nearly eliminates idling (5--9\% of the remaining failures \vs 69--96\% for the base planner) without penalizing any action.
On the four settings, random residuals of the same norm stay close to the base planner (supplementary material).

\noindent\textbf{Re-initialization.}
Without fine-tuning, re-initialization raises success the most on the four-setting average, followed by the memory alone, the stall penalty, restart at stalls, and uniform weights (\cref{tab:shape,tab:visual,tab:ablation}).
We combine re-initialization, the strongest of the three planning-time changes (restart at stalls, stall penalty, re-initialization), with the memory and fine-tuning.
With fine-tuning, no agent is idle in the remaining failures of \oursI, but the block still stays in place in 76--92\% of them.
As with CEM, removing the fixed point stops the agent from idling but leaves the block in place.
A random linear perturbation of the predicted latents gains as much as re-initialization if redrawn per replan but no more than fine-tuning alone if drawn per episode (supplementary material).

\noindent\textbf{Combinations.}
Fine-tuning adds to the memory and to re-initialization, since on the four-setting average each combination is above both of its parts.
The memory and re-initialization do not add to each other.
Together they stay close to re-initialization alone on average, with or without fine-tuning (\cref{tab:ablation}).

\subsection{Scope of the remedies}
\label{sec:scope}

\noindent\textbf{CEM.}
With CEM, fine-tuning alone stays close to \ours on the five settings where it was run (supplementary material), so the memory adds little.
The memory thus adds to fine-tuning mainly under GD, where the warm start creates the fixed point that it breaks.

\begin{table}[t]
\centering
\caption{\textbf{Other world models (PushT, GD).} DINO-WM: its own protocol, 20 MPC steps; other models: clean frames. DINOv2: frozen; proj.: global projector, excluded from fine-tuning; AdaJEPA: test-time weight updates. Mean $\pm$ SD over seeds. \textbf{Bold}, \underline{underline}: best, second best without test-time weight updates; ties share marks.}
\label{tab:models}
\small
\setlength{\tabcolsep}{2.6pt}
\begin{tabular}{l ccc}
\toprule
Success (\%) $\uparrow$ & \makecell{Main\\model} & \makecell{Temporal\\Straightening} & DINO-WM \\
Encoder & ResNet & DINOv2+proj. & DINOv2 \\
\midrule
Base                                      & 64.0\sd{1.6} & 30.7\sd{5.0} & \textbf{52.0}\sd{10.2} \\
\quad + \ours                             & \textbf{80.7}\sd{0.9} & \textbf{36.7}\sd{1.9} & \textbf{52.0}\sd{8.5} \\
\quad + \oursI                            & \underline{69.3}\sd{6.6} & \underline{31.3}\sd{6.2} & \underline{46.7}\sd{7.5} \\
\midrule
AdaJEPA~\cite{wang2026adajepa}  & 84.0\sd{1.6} & 28.7\sd{2.5} & 55.3\sd{3.4} \\
\bottomrule
\end{tabular}
\end{table}

\noindent\textbf{Other world models.}
On both models, 74--89\% of the base planner's failures are again stalls, though the agent mostly moves.
On Temporal Straightening, \ours raises success, while \oursI and adaptation change it little (\cref{tab:models}). On DINO-WM, \ours leaves it unchanged, and \oursI and adaptation change it within one SD.

\subsection{Failure cases}
\label{sec:failures}

\noindent\textbf{Red block.}
Under the red block and darkening at level 0.5, the failures differ from those in other settings (\cref{fig:analysis}a, bottom).
The agent keeps moving (at most 10\% idle), and the predicted progress is negative.
Linear probes~\cite{alain2017probes} locate the cause in the encoder.
The block position remains decodable, but the $R^2$ for the block angle drops from 0.95 on clean to 0.49 (red block) and 0.37 (dark 0.5).
The planning objective therefore largely ignores the angle, and there is no warm-start fixed point.
On the red block, \ours and \oursI leave success nearly unchanged (\cref{tab:visual}; \cref{fig:qual}f).

\noindent\textbf{PointMaze.}
The agent stalls although the model is accurate.
On the medium maze, 41--55\% of failures under default dynamics and high damping are idle (the agent moves less than 10\% of a typical successful step), and the model predicts the cost of staying almost exactly (median relative error $-1\%$).
At least 93\% of the idle failures are objective traps.
Simple plans move the agent closer to the goal, yet none lowers the objective even on simulated frames.
The agent stops at a local minimum of the planning objective, so the objective, not the model, is at fault.
There, the memory changes success by at most 5.3 points (\cref{tab:pointmaze}; \cref{fig:qual}h), and re-initialization lowers it under all three dynamics (\cref{fig:qual}i).
With fine-tuning, both remedies stay below the base planner on the medium maze, and on unseen layouts they fall from 62.0 (no fine-tuning) to 54.7, as selection-set success predicted.
On unseen layouts, whose idle failures resemble PushObj's, the memory and re-initialization each raise success from 57.3 to 62.0, uniform weights to 72.0, and adaptation to 64.0 and 76.7 (\cref{tab:pointmaze}; \cref{fig:qual}g).
A stall alone does not predict the remedies' benefit.
\begin{table}[t]
\centering
\caption{\textbf{Success under dynamics and layout shifts (\mbox{PointMaze}).} GD, 20 MPC steps; damp.: damping. \ours, \oursI: on-policy fine-tuning (\OFT) + residual memory (\RM), random re-initialization (\RI); \OFT rows: comparison only; AdaJEPA: encoder head + last/first predictor block adapted. Mean $\pm$ SD over seeds. \textbf{Bold}, \underline{underline}: best, second best excluding AdaJEPA; ties share marks.}
\label{tab:pointmaze}
\small
\setlength{\tabcolsep}{1.8pt}
\begin{tabular}{l cccc}
\toprule
Success (\%) $\uparrow$ & \multicolumn{3}{c}{Medium maze} & Layout \\
\cmidrule(lr){2-4}\cmidrule(lr){5-5}
Method & default & low mass & high damp. & unseen \\
\midrule
Base                                       & \underline{79.3}\sd{5.0} & \underline{76.7}\sd{5.0} & 68.7\sd{1.9} & 57.3\sd{8.2} \\
\quad + \RM                                & 78.0\sd{3.3} & \textbf{78.7}\sd{4.7} & \textbf{74.0}\sd{1.6} & \underline{62.0}\sd{4.3} \\
\quad + \RI                                & 70.7\sd{0.9} & 62.0\sd{1.6} & 55.3\sd{5.2} & \underline{62.0}\sd{9.1} \\
\quad + straight chunks                    & \textbf{80.7}\sd{0.9} & 66.7\sd{4.7} & 69.3\sd{4.1} & 60.7\sd{3.4} \\
\quad + uniform weights                    & 78.7\sd{2.5} & 76.0\sd{6.5} & \underline{72.0}\sd{1.6} & \textbf{72.0}\sd{5.9} \\
\quad + \OFT                               & 73.3\sd{5.2} & 72.0\sd{1.6} & 71.3\sd{4.7} & 49.3\sd{5.0} \\
\quad + \ours                              & 76.0\sd{6.5} & 74.7\sd{6.2} & 68.0\sd{3.3} & 54.7\sd{7.7} \\
\quad + \oursI                             & 77.3\sd{5.0} & 70.0\sd{8.5} & 60.0\sd{1.6} & 54.7\sd{2.5} \\
\midrule
AdaJEPA (last block)                       & 78.7\sd{3.4} & 76.0\sd{4.3} & 70.0\sd{4.3} & 64.0\sd{7.1} \\
AdaJEPA (first block)                      & 81.3\sd{4.1} & 72.7\sd{5.7} & 74.0\sd{3.3} & 76.7\sd{5.2} \\
\bottomrule
\end{tabular}
\end{table}

\section{Conclusion}
\label{sec:conclusion}
We propose Unstick, which breaks the planning-loop fixed point behind most failures of JEPA planners on pushing tasks under GD, with fixed weights.
These failures are stalls, where the object stays in place while the model predicts progress, with and without shift.
At the analyzed stalls, the model is miscalibrated on its own actions, and warm-started GD then repeats nearly the same plan.
Unstick recovers most of the adaptation gain with the memory, and with re-initialization most of it on shape but much less on visual shifts, yet little where the encoder or objective is at fault.
Whether these planners succeed under shift thus depends on the planning loop, not only the model.

\paragraph{Limitations.}
The stall diagnosis covers one JEPA family on three benchmarks, mainly pushing tasks planned with GD, with fewer idle failures on PointMaze.
Stalls recur on two other JEPA models, and testing reconstruction-based world models would show whether the fixed point also limits them.
Both remedies target the fixed point of warm-started GD, and with CEM the memory adds little to fine-tuning.
The memory corrects predictions only near transitions the episode has visited and grows with every chunk, so long episodes require a bounded memory.
Fine-tuning requires base-planner executions on episodes separate from the evaluation episodes before deployment.

\section*{LLM Usage Statement}
A large language model (Claude, Anthropic), used through Claude Code, assisted in writing and editing the text of this paper and its supplementary material, including checking the text against the venue's formatting and style guidelines. It also generated revisions of the plotting scripts for the figures. The authors reviewed all edits and take full responsibility for the content.

{
    \small
    \setlength{\bibsep}{0pt}
    \bibliographystyle{ieeenat_fullname}
    \bibliography{main}
}

\end{document}